\documentclass{article}
\usepackage{graphicx}

   \usepackage[preprint,nonatbib]{neurips_2019}

\usepackage[utf8]{inputenc} % allow utf-8 input
\usepackage[T1]{fontenc}    % use 8-bit T1 fonts
\usepackage{hyperref}       % hyperlinks
\usepackage{url}            % simple URL typesetting
\usepackage{booktabs}       % professional-quality tables
\usepackage{amsfonts}       % blackboard math symbols
\usepackage{nicefrac}       % compact symbols for 1/2, etc.
\usepackage{microtype}      % microtypography

\title{Generative Adversarial Networks are Special Cases of Artificial Curiosity (1990) and also Closely Related to  Predictability Minimization (1991)}

\author{J\"{u}rgen Schmidhuber  \\ The Swiss AI Lab, IDSIA, USI \& SUPSI, Manno-Lugano \\ NNAISENSE, Lugano, Switzerland \\
Preprint (accepted by Neural Networks, Volume 127, July 2020, Pages 58-66) %https://doi.org/10.1016/j.neunet.2020.04.008
}

\begin{document}

\maketitle

\begin{abstract}

I review unsupervised or self-supervised neural networks playing minimax games in game-theoretic settings: (i) Artificial Curiosity (AC, 1990) is based on two such networks. One network learns to generate a probability distribution over outputs, the other learns to predict effects of the outputs. Each network minimizes the objective function maximized by the other. (ii) Generative Adversarial Networks (GANs, 2010-2014) are an application of AC where the effect of an output is 1 if the output is in a given set, and 0 otherwise. (iii) Predictability Minimization (PM, 1990s) models data distributions through a neural encoder that maximizes the objective function minimized by a neural predictor of the code components.  I correct a previously published claim that PM is not based on a minimax game.

\end{abstract}

\section{Introduction}
\label{intro}

Computer science has a rich history of problem solving through computational procedures seeking to minimize an objective function maximized by another procedure. For example, chess programs date back to 1945~\cite{zuse45chess}, and for many decades have successfully used a recursive minimax procedure with continually shrinking look-ahead, e.g.,~\cite{wiener1965}.
Game theory of adversarial players originated in 1944~\cite{morgenstern1944}.
In the field of machine learning, early adversarial settings include reinforcement learners playing against themselves~\cite{Samuel:59} (1959), or the evolution of parasites in predator-prey games, e.g.,~\cite{hillis1990co,seger1988parasites} (1990).

In 1990, a new type of adversarial technique
was introduced in the field of  {\em  unsupervised or self-supervised artificial neural networks} (NNs)~\cite{Schmidhuber:90diffenglish,Schmidhuber:90sab} (Sec. \ref{ac}).  Here a  {\em single} agent has two separate learning NNs.  Without a teacher, and without external reward for achieving user-defined goals, the first NN somehow generates outputs. The second NN learns to  predict consequences or properties of the generated outputs, minimizing its errors, typically by gradient descent. However, the first NN  {\em  maximizes} the objective function {\em  minimized} by the second NN, effectively trying to generate data from which the second NN can  still learn
to improve its predictions. 

This survey will review such  {\em unsupervised minimax} techniques, and relate them to each other. Sec. \ref{ac} focuses on  unsupervised Reinforcement Learning (RL)  through {\em Artificial Curiosity} (since 1990). Here the prediction errors  are (intrinsic) reward signals maximized by an RL controller. Sec. \ref{special}  
points out that {\em Generative Adversarial Networks} (GANs, 2010-2014) and its variants are special cases of this approach. Sec. \ref{brains} discusses a more sophisticated adversarial approach of 1997. Sec. \ref{pm} 
addresses unsupervised encoding of data through {\em Predictability Minimization} (PM, 1991), where the predictor's error is maximized by the encoder's feature extractors. Sec. \ref{convergence} addresses issues of convergence.

For historical accuracy, I will sometimes refer not only to peer-reviewed publications but also to technical reports, many of which turned into reviewed journal articles later.

\newpage
\section{Adversarial Artificial Curiosity (AC, 1990)}
\label{ac}

In 1990, unsupervised or self-supervised adversarial NNs  were
used to implement  {\em  curiosity}~\cite{Schmidhuber:90diffenglish,Schmidhuber:90sab} 
in the general context of exploration in 
RL~\cite{Kaelbling:96,Sutton:98,wiering2012}
(see Sec. 6 of~\cite{888} for a survey of deep RL).
The goal was to overcome drawbacks of traditional reward-maximizing RL machines 
which use naive strategies (such as random action selection)  to explore their environments.  

The basic idea is: An RL agent with a predictive NN world model maximizes intrinsic reward obtained for 
provoking situations where the error-minimizing world model still has high error 
and can learn more.
I will refer to this approach as {\em  Adversarial Artificial Curiosity} (AC) of 1990, or 
AC1990 for short, to distinguish it from our later types of 
Artificial Curiosity since 1991 (Sec. \ref{ac1991}). 

In what follows, let $m,n,q$ denote positive integer constants.
In the AC context, the first NN is often called the controller C. 
C may interact with an environment through 
sequences of interactions  called  {\em trials} or {\em episodes}. 
During the execution of  a single interaction of any given trial,
C generates an output vector $x \in \mathbb{R}^n$. This may influence 
an environment, which produces a reaction to $x$ in form of an 
observation $y \in \mathbb{R}^q$.
In turn, $y$ may affect C's inputs during the next interaction if there is any.

In the first variant of AC1990~\cite{Schmidhuber:90diffenglish,Schmidhuber:90sab}, 
C is recurrent, and thus a general purpose computer. 
Some of C's  adaptive recurrent units are 
 mean and variance-generating Gaussian units, 
 such that C can become a {\em generative model} that produces a  probability distribution over outputs---see 
Section {\em ``Explicit Random Actions versus Imported Randomness''}~\cite{Schmidhuber:90diffenglish} (see also~\cite{Schmidhuber:91nips,Williams:88b}).
(What these stochastic units
 do can be equivalently accomplished by having C perceive pseudorandom numbers or noise, 
like the generator NNs of GANs~\cite{goodfellow2014generative}; Sec. \ref{special}).

To compute an output action during an interaction, C updates all its NN unit activations 
for several discrete time steps in a row---see Section {\em ``More Network Ticks than Environmental Ticks''}~\cite{Schmidhuber:90diffenglish}. 
In principle, this allows for computing highly nonlinear, stochastic mappings from 
environmental inputs  (if there are any) and/or from internal ``noise'' to outputs.

The second NN is called the world model M~\cite{Schmidhuber:90diffenglish,Schmidhuber:90sandiego,Schmidhuber:91nips,ha2018world}.
In the first variant of  AC1990~\cite{Schmidhuber:90diffenglish,Schmidhuber:90sab}, M is also recurrent, for reasons of generality. 
M receives C's  outputs $x \in \mathbb{R}^n$  as inputs and predicts their visible 
environmental effects or consequences
$y \in \mathbb{R}^q$. 

According to AC1990, M  {\em minimizes}  its prediction errors by gradient descent, thus becoming a better predictor. In absence of external reward, however, the adversarial C tries to find actions that  {\em maximize}  the errors of M:  {\em  M's errors are the intrinsic rewards of C.}  Hence C {\em maximizes}  
the errors that M {\em minimizes}. 
The loss of M is the gain of C.

Without external reward, C is thus intrinsically motivated to invent novel action sequences or experiments that yield data that M still finds surprising, until the data becomes familiar and boring. 

 The 1990 paper~\cite{Schmidhuber:90diffenglish} 
describes gradient-based learning methods for both C and M.
In particular, {\em backpropagation~\cite{Linnainmaa:1970,Linnainmaa:1976}  through the model M down into the controller C} (whose outputs are  inputs to M) is used to compute weight changes for C, generalizing  previous work on feedforward networks~\cite{Werbos:89identification,Werbos:87specifications,Munro:87,JordanRumelhart:90}.
This is closely related to how 
the code generator NN of {\em Predictability Minimization} (Sec. \ref{pm}) can be trained by backpropagation through its predictor NN~\cite{Schmidhuber:91predmin,Schmidhuber:92ncfactorial,Schmidhuber:96ncedges},
and to how
the GAN generator NN (Sec. \ref{special}) can be trained by backpropagation through its discriminator NN~\cite{olli2010,goodfellow2014generative}.
Furthermore,  the concept of {\em backpropagation through random number generators}~\cite{Williams:88b} is used
 to derive error signals even
 for those units of C that are stochastic~\cite{Schmidhuber:90diffenglish}.

However, the original AC1990 paper 
points out that the basic ideas of AC are not limited to particular learning algorithms---see Section {\em ``Implementing Dynamic Curiosity and Boredom''}~\cite{Schmidhuber:90diffenglish}. 
Compare more recent summaries and later variants / extensions of AC1990's simple but powerful exploration principle~\cite{Schmidhuber:06cs,Schmidhuber:10ieeetamd}, which inspired much later work, e.g.,~\cite{Singh:05nips,Oudeyer:12intrinsic,Schmidhuber:10ieeetamd}; compare~\cite{oudeyerkaplan07,pathak2017curiosity,burda2018curious}.
See also related work of 1993~\cite{slg1993,slg2019}.

To summarize, unsupervised or self-supervised minimax-based neural networks of the previous millennium 
(now often called CM systems~\cite{learningtothink2015}) were both {\em adversarial} and {\em generative} (using terminology of 2014~\cite{goodfellow2014generative}, Sec. \ref{special}), stochastically generating outputs yielding experimental data, not only for stationary patterns but also for pattern sequences, even for the general case of RL, and even for {\em recurrent} NN-based RL in partially observable environments~\cite{Schmidhuber:90diffenglish,Schmidhuber:90sab}.

\newpage
\section{A Special Case of AC1990: Generative Adversarial Networks}
\label{special}

Let us now consider a special case of a curious CM system as in  Sec. \ref{ac} above, where each sequence of interactions  of the CM system with its environment (each trial) is limited to a  {\em single} interaction, like in bandit problems~\cite{robbins1952bandit,Gittins:89,auer1995,audibert2009minimax}. 

The environment contains a representation of a user-given training set $X$ of patterns $ \in \mathbb{R}^n$. $X$ is not directly visible to C and M, but its properties are  probed by AC1990 through C's outputs or actions or experiments.

In the beginning of any given trial, the activations of all units in C are reset. 
C is blind (there is no input from the environment).  
Using its internal stochastic units~\cite{Schmidhuber:90diffenglish,Schmidhuber:91nips} (Sec. \ref{ac}),  C then computes a single output $x \in \mathbb{R}^n$. In a pre-wired fraction of all cases,  $x$ is replaced by a randomly selected ``real'' pattern $\in X$ 
(the simple default exploration policy of traditional RL chooses a random action in a fixed fraction of all cases~\cite{Kaelbling:96,Sutton:98,wiering2012}). 
This ensures that M will see both ``fake'' and ``real'' patterns. 

The environment will react to output  $x$ and return as its effect a binary observation $y \in \mathbb{R}$, where  $y=1$ if $x \in X$, and $y=0$ otherwise. 

As always in AC1990-like systems, M now takes C's output $x$ as an input, and predicts its environmental effect $y$, in that case a single bit of information, 1 or 0. As always, M learns by {\em minimizing}  its prediction errors. However, as always in absence of external reward, the adversarial C is learning to generate outputs that  {\em maximize}  the error  {\em minimized} by M. M's loss is C's negative loss. 
Since the stochastic C is trained to {\em maximize} 
 the objective function {\em  minimized} by M, C is 
motivated  to produce a distribution over more and more realistic patterns, e.g., images. 

Since 2014, this particular application of the AC principle (1990) has been called a {\em Generative Adversarial Network} (GAN)~\cite{goodfellow2014generative}. M was called 
 the discriminator,  
 C was called  the generator.
GANs and related approaches 
are now widely used and studied, 
e.g.,~\cite{radford2015,denton2015,huszar2015not,nowozin2016f,wasserstein2017,ganin2016,makhzani2015,bousmalis2016,underthiner2017}.

\subsection{Additional comments on AC1990 \& GANs \& Actor-Critic}
\label{comments}

The first variant of AC1990~\cite{Schmidhuber:90diffenglish,Schmidhuber:90sab,reddit2019gan} 
generalized to the case of recurrent NNs  a well-known way~\cite{Werbos:89identification,Werbos:87specifications,Munro:87,NguyenWidrow:89,JordanRumelhart:90,SchmidhuberHuber:91}  of using a differentiable world model M to approximate gradients for C's parameters even when environmental rewards are  {\em non-differentiable} functions of C's actions. In the simple differentiable GAN environment above, however, there are no such complications, since the rewards of C (the 1-dimensional errors of M) are differentiable functions of C's outputs. That is, standard backpropagation~\cite{Linnainmaa:1970} can directly compute the gradients of C's parameters with respect to C's rewards, like in Predictability Minimization (1991)~\cite{Schmidhuber:91predmin,Schmidhuber:92ncfactorial,schmidhuber1993,Schmidhuber:96ncedges,pm,Schmidhuber:99zif} (Sec. \ref{pm}). 

Unlike the first variant of AC1990~\cite{Schmidhuber:90diffenglish,Schmidhuber:90sab}, most current GAN applications use more limited 
 feedforward NNs rather than recurrent NNs to implement M and C. The stochastic units of C are typically implemented by feeding noise sampled from a given probability distribution into C's inputs~\cite{goodfellow2014generative}.\footnote{In the GAN-like AC1990 setup of Sec. \ref{special},  
real patterns (say, images) are produced in a pre-wired fraction of all cases. However,
  one could easily give C the freedom to decide by itself to focus on particular  {\em real} images $\in X$ that  M finds still difficult to process. For example, one could employ the following procedure: once C has generated a fake image $\hat{x} \in \mathbb{R}^n$, and the activation of a special hidden unit of C is above a given threshold, say, 0.5, then $\hat{x}$ is replaced by the pattern in $X$ most similar to $\hat{x}$, according to some similarity measure.  In this case, C is not only motivated to learn to generate almost realistic
 {\em fake}   images that are still hard to classify by M, but also to address and focus on those {\em real} images that are still hard on M. This may be useful as C sometimes may find it easier to fool M by sending it a particular real image, rather than a fake image.  To my knowledge, however, this is rarely done with standard GANs.}
 
Actor-Critic methods~\cite{konda2000actor,Sutton:99} are less closely related to GANs as they do not embody explicit minimax games. Nevertheless, a GAN can be seen as a  {\em modified} Actor-Critic with a blind C in a stateless MDP~\cite{pfau2016connect}. This in turn yields another connection between AC1990 and Actor-Critic (compare also Section 
 {\em "Augmenting the Algorithm by Temporal Difference Methods”}~\cite{Schmidhuber:90diffenglish}).

 \newpage
\subsection{A closely related special case of AC1990: Conditional GANs (2010)}
\label{cgan}

Unlike AC1990~\cite{Schmidhuber:90diffenglish} and the GAN of 2014~\cite{goodfellow2014generative}, the GAN of 2010~\cite{olli2010} 
(now known as a  {\em conditional} GAN or cGAN~\cite{mirza2014conditional}) does  {\em not} have an internal source of randomness. Instead, such cGANs depend on sufficiently diverse inputs from the environment. 

cGANs are also  special cases of the AC principle (1990):
 cGAN-like additional environmental inputs just mean that the controller C of AC1990  is not blind any more
like in the example above with the GAN of 2014~\cite{goodfellow2014generative}. 

Like the first version of AC1990~\cite{Schmidhuber:90diffenglish}, the cGAN of 2010~\cite{olli2010}  minimaxed {\em  Least Squares} errors. This was later called LSGAN~\cite{mao2017least}. 

\subsection{AC1990 and StyleGANs (2019)}
\label{style}

The GAN of 2014~\cite{goodfellow2014generative} perceives noise vectors (typically sampled from a Gaussian) in its input layer  and maps them to outputs. The more general StyleGAN~\cite{karras2019style}, however, allows for noise injection in deeper hidden layers as well, to implement all sorts of hierarchically structured probability distributions. 

Note that this kind of additional probabilistic 
expressiveness was already present in the mean and variance-generating Gaussian units of the recurrent generator network C of 
AC1990~\cite{Schmidhuber:90diffenglish} (Sec. \ref{ac}).

\subsection{Summary: GANs and cGANs etc. are simple instances of AC1990}
\label{summary}

cGANs (2010) and GANs (2014) are quite different from certain
earlier adversarial machine learning settings~\cite{Samuel:59,hillis1990co} (1959-1990) 
which 
neither  involved unsupervised neural networks nor 
were about modeling  data nor used gradient descent  (see Sec. \ref{intro}). 
However, GANs and cGANs  are very closely related to AC1990.

GANs are essentially an application of the Adversarial Artificial Curiosity principle of 1990 (Sec. \ref{ac}) where the generator network C is blind and
the environment simply returns whether C's current output is in a given set. 
 As always, C maximizes the function minimized by M (Sec. \ref{special}).

Same for cGANS, except that in this case C is not blind any more (Sec. \ref{cgan}). 

Similar for StyleGANs (Sec. \ref{style}).

\subsection{The generality of AC1990}
\label{general}

It should be emphasized though that AC1990 has much broader applicability~\cite{Singh:05nips,Oudeyer:12intrinsic,Schmidhuber:10ieeetamd,burda2018curious} 
than the GAN-like special cases above. In particular, C may sequentially interact with the environment for a long time, producing a sequence of environment-manipulating outputs resulting in complex environmental constructs. For example, C may trigger actions that generate brush strokes on a canvas, incrementally refining a painting over 
time, e.g.,~\cite{ha2017sketch,ganin2018synth,zheng2018stroke,huang2019paint,nakano2019paint}. Similarly, M may sequentially predict many other aspects of the environment besides the single bit of information in the GAN-like setup above. 
General AC1990 is about unsupervised or self-supervised RL agents that actively shape their observation streams through their own actions, setting themselves their own goals through intrinsic rewards, exploring the world by inventing their own action sequences or experiments, to discover novel, previously unknown predictability in the data generated by the experiments. 

Not only the 1990s but also recent years saw successful applications of this simple principle (and variants thereof) in sequential settings, e.g.,~\cite{pathak2017curiosity,burda2018curious}.

Since the GAN-like environment above is  restricted to a teacher-given set $X$ of patterns and a procedure deciding whether a given pattern is in $X$, the teacher will find it rather easy to evaluate the quality of C's $X$-imitating behavior. In this sense the GAN setting is ``more'' supervised than certain other applications of AC1990, which may be ``highly" unsupervised in the sense that  C may have much more freedom when it comes to selecting environment-affecting actions.

%\newpage
\section{Improvements of AC1990}
\label{ac1991}

Numerous improvements of the original AC1990~\cite{Schmidhuber:90diffenglish,Schmidhuber:90sab} are summarized in more recent surveys~\cite{Schmidhuber:06cs,Schmidhuber:10ieeetamd}. Let us focus here on a first important improvement of 1991. 

The errors of AC1990's M (to be {\em minimized}) are the rewards of its C (to be {\em maximized}, Sec. \ref{ac}). 
This makes for a fine exploration strategy in many deterministic environments.
In stochastic environments, however, this might fail. 
C might learn to  focus on those 
parts of the environment where M can always 
get high prediction errors due to randomness, 
or due to computational limitations of M. 
For example,  an agent controlled by C might get stuck in front of 
a  TV screen  showing highly unpredictable white  
noise, e.g.,~\cite{Schmidhuber:10ieeetamd} (see also~\cite{burda2018curious}).

Therefore, as pointed out in 1991,
in stochastic environments, 
C's reward  should not be the errors of M, 
but (an approximation of) the {\em first derivative} of M's errors across subsequent training iterations,
that is, M's {\em improvements}~\cite{Schmidhuber:91singaporecur,Schmidhuber:07alt}. 
As a consequence, despite M's high errors in front of 
the noisy TV screen above,
C won't get rewarded for getting stuck there,
simply because M's errors won't improve.
Both the totally predictable  and the fundamentally unpredictable  will get boring. 

This insight led to lots of follow-up work~\cite{Schmidhuber:10ieeetamd}. 
For example, 
one particular RL approach for AC in stochastic environments was published in 1995~\cite{Storck:95}. 
A simple M learned to predict or estimate the probabilities of the 
environment's possible responses, given C's actions. 
After each interaction with the environment,
C's reward was the KL-Divergence~\cite{kullback1951} 
between M's estimated probability distributions
before and after the resulting new experience (the information gain)~\cite{Storck:95}.
(This was later also called {\em Bayesian Surprise}~\cite{itti:05}; 
compare earlier work on information gain 
and its maximization {\em without} NNs~\cite{Shannon:48,Fedorov:72}.)

AC1990's above-mentioned limitations in probabilistic environments, 
however, are not an issue  in
the simple GAN-like setup of Sec. \ref{special},
because there the environmental reactions are totally deterministic:
For each image-generating action of C, 
there is a unique deterministic binary response from the environment
stating whether the generated image is in $X$ or not. 

Hence it is not obvious that above-mentioned improvements of AC1990 
hold promise also for GANs.

\section{AC1997: Adversarial Brains Bet on Outcomes of Probabilistic Programs}
\label{brains}

Of particular interest in the context of the present paper is
one more advanced adversarial approach to curious exploration of 1997~\cite{Schmidhuber:97interesting,Schmidhuber:99cec,Schmidhuber:02predictable}, referred to as AC1997.
AC1997 is about generating computational experiments in form of programs whose execution may change both an external environment and the RL agent's internal state. An experiment has a binary outcome: either a particular effect happens, or it doesn't. Experiments are collectively proposed by two reward-maximizing adversarial policies. Both can predict and bet on experimental outcomes before they happen. Once such an outcome is actually observed,  the  winner will get a positive reward proportional to the bet, and the loser a negative reward of equal magnitude. So each policy is motivated to create experiments whose yes/no outcomes surprise the other policy. The latter in turn is motivated to learn something about the world that it did not yet know, such that it is not outwitted again.

 More precisely, a single RL agent has two dueling, 
 reward-maximizing  {\em policies} called the 
 {\em  left brain} and the {\em right brain}.
Each brain is a modifiable probability distribution over programs 
running on a general purpose computer.
 {\em Experiments} are programs sampled in a collaborative way that is influenced by both brains.  
 Each experiment specifies how to execute an instruction  sequence (which may affect both the environment and the agent's internal state), and how to compute the outcome of the experiment through instructions implementing a computable function (possibly resulting in an internal binary yes/no classification) of the observation sequence triggered by the experiment.
 The modifiable parameters of
both brains are instruction probabilities. They 
can be accessed and manipulated through programs that include
 subsequences of special {\em self-referential}  policy-modifying instructions~\cite{Schmidhuber:94self,Schmidhuber:97ssa}.  
 
 Both brains may also 
trigger the execution of certain  {\em bet} instructions whose effect is to predict experimental outcomes before they are observed. If their predictions or hypotheses differ, they may agree to execute the experiment to determine which brain was right, and the surprised loser will pay an intrinsic reward (the real-valued bet, e.g., 1.0) to the winner in a zero sum game. 

That is, each brain is intrinsically motivated to 
 outwit or surprise the other by proposing an experiment such 
 that the other {\em agrees} on the experimental 
 protocol but {\em disagrees} on the predicted outcome.
 This outcome is typically an internal computable abstraction 
of complex spatio-temporal events generated through 
the execution the self-invented experiment.

This motivates the unsupervised or self-supervised two brain system 
to focus on  "interesting" computational questions, 
losing interest in "boring"
computations (potentially involving the environment) 
whose outcomes are consistently predictable by {\em both} brains, 
as well as computations whose outcomes are currently still hard to predict by {\em either} brain. 
Again, in the absence of external reward, 
each brain maximizes the value function minimised by the other. 

Using the  meta-learning  {\em  Success-Story RL algorithm}~\cite{Schmidhuber:94self,Schmidhuber:97ssa}, AC1997 learns when to learn and what to learn~\cite{Schmidhuber:97interesting,Schmidhuber:99cec,Schmidhuber:02predictable}. AC1997 
will also minimize the computational cost of learning new skills,
provided both brains receive a small 
negative reward for each computational step, which
introduces a bias towards  {\em simple} still surprising experiments (reflecting {\em simple} still unsolved problems). This may facilitate hierarchical construction of more and more complex experiments, including those yielding  {\em external} reward (if there is any). In fact, AC1997's artificial  creativity may not only drive artificial scientists and artists, e.g.,~\cite{Schmidhuber:09multiple}, but can also accelerate the intake of external reward, e.g.,~\cite{Schmidhuber:97interesting,Schmidhuber:02predictable}, intuitively because a better understanding of the world can help to solve certain problems faster. 

Other RL or evolutionary  algorithms could also be applied to such 
two-brain systems implemented as two interacting (possibly recurrent) RL  NNs or other computers. 
However, certain issues such as 
catastrophic forgetting are presumably better addressed by the later 
{\sc PowerPlay} framework (2011)~\cite{powerplay2011and13,Srivastava2013first},
which offers an  {\em  asymptotically optimal} way of finding the simplest yet unsolved problem 
in a (potentially infinite) set of formalizable problems with computable solutions,
and adding its solution to the repertoire of a more and more general, curious problem solver. 
Compare also the {\em One Big Net For Everything}~\cite{onebignet2018} which offers a simplified, less strict NN 
version of {\sc PowerPlay}. 

How does AC1997 relate to GANs? AC1997 is similar to standard GANs in the sense that both are unsupervised  generative adversarial minimax players and focus on experiments with a binary outcome: {\em  1 or 0, yes or no, hypothesis true or false.} However, for GANs the experimental protocol is prewired and always the same: It simply tests whether a recently generated pattern is in a given set or not (Sec. \ref{special}). One can restrict AC1997 to such simple settings 
by limiting its domain and the nature of the instructions in its programming language, such that  possible bets of both brains are limited to binary yes/no outcomes of GAN-like experiments. 
In general, however, the adversarial brains of AC1997 can invent essentially arbitrary computational questions or problems by themselves, generating programs that interact with the environment in any computable way that will yield binary results on which both brains can bet. 
A bit like a pure scientist deriving internal joy signals from inventing experiments that yield discoveries of initially surprising but learnable and then reliably repeatable predictabilities. 

%\vskip -24pt
%\newpage
\section{Predictability Minimization (PM)}
\label{pm}

An important NN task is to learn the statistics of given data such as images. To achieve this, the principles of gradient descent/ascent were used in {\em yet another type of unsupervised minimax game} where one NN minimizes the objective function maximized by another. This duel between two unsupervised adversarial NNs was introduced in the 1990s in a series of papers~\cite{Schmidhuber:91predmin,Schmidhuber:92ncfactorial,schmidhuber1993,Schmidhuber:96ncedges,pm,Schmidhuber:99zif}. It   was called {\em Predictability Minimization (PM)}. 

PM's goal is to achieve an important goal of unsupervised learning, namely, 
an ideal, disentangled, {\em factorial} code~\cite{Barlow:89,Barlow:89review}
of given data, where the code components are statistically independent of each other.
That is, {\em  the codes are distributed like the data, and   the probability of a given data pattern is simply the product of the probabilities of its code components.} 
%Hence for many classification tasks a {\em naive} Bayes classifier can subsequently take the code and classify the data both efficiently and optimally~\cite{Schmidhuber:96ncedges,pm}, to the extent that  its naive assumption of conditionally independent variables is actually {\em valid}.
Such codes may facilitate subsequent downstream learning~\cite{Schmidhuber:96ncedges,pm,Schmidhuber:99zif}. 

PM requires an encoder network with initially random weights. 
It maps data samples $x \in \mathbb{R}^n$ (such as images)
to codes  $y \in [0,1]^m$
represented across $m$ so-called code units. 
In what follows, integer indices $i,j$ range over $1,\ldots,m$.
The $i$-th component of $y$ is called $y_i \in [0,1]$.
A separate predictor network is trained by gradient descent to predict each $y_i$ from the remaining components $y_j (j \neq i)$.
The encoder, however, is trained to maximize the same objective function (e.g., mean squared error) 
minimized by the predictor. Compare  
the text near Equation 2 in the 1996 paper~\cite{Schmidhuber:96ncedges}: {\em``The clue is:  the code units are trained (in our experiments by online backprop) to maximize essentially the same objective function the predictors try to minimize;"} or Equation 3 in  Sec. 4.1 of the 1999 paper~\cite{Schmidhuber:99zif}: {\em ``But the code units try to maximize the same objective function the predictors try to minimize."} 
%See also  Appendix~\ref{app} for formal details. 

Why should the end result of this fight between predictor and encoder be a
disentangled  factorial code?
Using gradient descent, to maximize the prediction errors, 
the code unit activations $y_j$ run away from their real-valued predictions in $[0,1]$, that is, 
they are forced towards the corners of the unit interval, and tend to become binary, either 0 or 1.
And according to a proof of 1992~\cite{DayanZemel:92,schmidhuber1993},\footnote{It should be mentioned that the above-mentioned proof~\cite{DayanZemel:92,schmidhuber1993} is limited to binary factorial codes. There is no proof that PM is a universal method for approximating all kinds of non-binary distributions (most of which are incomputable anyway). Nevertheless, it is well-known that binary Bernoulli distributions can approximate at least Gaussians and other distributions, that is, with enough binary code units one should get at least arbitrarily close approximations of broad classes of distributions. In the PM papers of the 1990s, however, this was not studied in detail.}
%by Dayan \& Zemel \& Pouget
the encoder's objective function is maximized when the $i$-th code unit maximizes its variance
(thus maximizing the information it conveys about the input data) 
while simultaneously minimizing the deviation between its (unconditional) expected activations $E(y_i)$
and its predictor-modeled, {\em conditional} expected activations $E(y_i \mid \{y_j, j \neq i \})$, given the other code units. 
See also conjecture 6.4.1 and  Sec. 6.9.3 of the thesis~\cite{schmidhuber1993}.
%and Appendix~\ref{app}.
That is, the code units are motivated to extract informative yet mutually independent binary features from the data.

PM's inherent class of probability distributions is the set 
of {\em multivariate binomial distributions.}
In the ideal case, PM has indeed learned to create a binary factorial code of the data.
That is, 
in response to some input pattern, 
each $y_i$ is either 0 or 1, 
and the predictor has learned the conditional expected value $E(y_i \mid \{y_j, j \neq i \})$.
Since the code is both binary and factorial,
this value is equal to the code unit's {\em unconditional} probability $P(y_i=1)$ of being on
(e.g., ~\cite{Schmidhuber:92ncfactorial}, Equation in Sec. 2).
E.g., if some code unit's prediction is 0.25, 
then the probability of this code unit being on is 1/4. 

The first toy experiments with PM~\cite{Schmidhuber:91predmin} were 
conducted nearly three decades ago when compute was about a million times more expensive than today.
When it had become about 10 times cheaper 5 years later, 
it was shown that simple semi-linear PM variants applied to images automatically generate  
feature detectors well-known from neuroscience, such as 
on-center-off-surround detectors,
off-center-on-surround detectors,
orientation-sensitive bar detectors, etc~\cite{Schmidhuber:96ncedges,pm}. 

%\vskip -24pt
\subsection{Is it true that PM is NOT a minimax game?}
\label{true}

The NIPS 2014 GAN paper~\cite{goodfellow2014generative}
states that PM differs from GANs in the sense that PM is NOT based on a minimax game with a value function that one agent seeks to maximize and the other seeks to minimise. It states that for GANs 
{\em "the competition between the networks is the sole training criterion, and is sufficient on its own to train the network,"} while PM {\em "is only a regularizer that encourages the hidden units of a neural network to be statistically independent while they accomplish some other task; it is not a primary training criterion"}~\cite{goodfellow2014generative}. 
%This claim has also obtained a certain prominence in social networks. 

But this claim is incorrect, 
since PM is indeed a pure minimax game, too, e.g.,~\cite{Schmidhuber:96ncedges}, Equation 2. 
There is no {\em "other task."} In particular, PM was also trained ~\cite{Schmidhuber:91predmin,Schmidhuber:92ncfactorial,schmidhuber1993,Schmidhuber:96ncedges,pm,Schmidhuber:99zif} (also on images ~\cite{Schmidhuber:96ncedges,pm}) such that {\em "the competition between the networks is the sole training criterion, and is sufficient on its own to train the network."}

%\vskip -24pt
\subsection{Learning generative models through PM variants}
\label{pmgen}
%(not done in previous PM work)

One of the variants in the first peer-reviewed PM paper (\cite{Schmidhuber:92ncfactorial} e.g., Sec 4.3, 4.4) had an optional decoder (called {\em reconstructor}) attached 
 to the code such that data can be reconstructed from its code.
%(compare Appendix~\ref{app}).
% (i.e., $\beta \neq 0$ in Equation 4).
 Let's assume that PM has indeed found an ideal  factorial code of the data.
 Since the codes are distributed like the data,
with the decoder, 
we could immediately use the system as a {\em generative model,}
by randomly activating each binary code unit according to its unconditional probability 
(which for all training patterns is now equal to the activation of its prediction---see  Sec. \ref{pm}),
and sampling output data through the decoder.\footnote{Note that even one-dimensional data may have a  complex distribution whose binary factorial code (Sec. \ref{pm}) may require many dimensions. PM's goal is the discovery of such a  code, with an a priori unknown number of components. For example, if there are 8 input patterns, each represented by a single real-valued number between 0 and 1, each occurring with probability 1/8, then there is an ideal binary factorial code across 3 binary code units, each active with probability 1/2. Through a decoder on top of the 3-dimensional code of the  1-dimensional data we could resample the original data distribution, by randomly activating each of the 3 binary code units with probability 50\% (these probabilities are actually directly visible as predictor activations).}
With an accurate decoder, 
 the sampled data must obey the statistics of the original distribution,
by definition of factorial codes.

However, to my knowledge,
this straight-forward application as a generative model was never explicitly mentioned in any 
PM paper, and  the decoder 
(as well as additional, optional local variance maximization for the code units)
was actually omitted in several PM papers after 
1993~\cite{Schmidhuber:96ncedges,pm,Schmidhuber:99zif}
which focused on unsupervised learning of disentangled internal representations,
to facilitate subsequent downstream learning~\cite{Schmidhuber:96ncedges,pm,Schmidhuber:99zif}. 

Nevertheless, generative models producing data 
through stochastic outputs of minimax-trained NNs  
were described in 1990~\cite{Schmidhuber:90diffenglish,Schmidhuber:90sab} (see Sec. \ref{ac} on Adversarial Artificial Curiosity)
and
2014~\cite{goodfellow2014generative}  (Sec. \ref{special}).
Compare also the concept of Adversarial Autoencoders~\cite{makhzani2015}.

\subsection{Learning factorial codes through GAN variants}
\label{ganfac}

PM variants could easily be used as GAN-like generative models (Sec. \ref{pmgen}).  In turn, GAN variants could easily be used to learn factorial codes like PM. 
If we take a GAN generator network trained on random input codes with independent components,
and attach a traditional encoder network to its output layer, and train this encoder to map the output patterns back to their original random codes,
then in the ideal case this encoder will become a factorial code generator that can also be applied to the original data. This was not done by the GANs of 2014~\cite{goodfellow2014generative}. However, compare  InfoGANs~\cite{infogan2016} and related 
work~\cite{makhzani2015,donahue2016adversarial,dumoulin2016adversarial}.

\subsection{Relation between PM and GANs and their variants}
\label{pmgan}

Both PM and GANs are unsupervised learning techniques
that  model the statistics of given data.
Both employ  gradient-based adversarial nets that play a minimax game to achieve their goals.

While PM tries to make easily decoded, random-looking, factorial codes of the data, 
GANs try to make decoded data directly from random codes. 
In this sense, 
the inputs of PM's encoders are like the outputs of GAN's  decoders,
while the outputs of PM's encoders are like the inputs of GAN's decoders.
In another sense, the outputs of PM's encoders are like the outputs of GAN's decoders because both are shaped by the  adversarial loss. 

Effectively, GANs are trying to approximate the true data distribution through 
some other distribution of a given type (e.g. Gaussian, binomial, etc). 
Likewise, PM is trying to approximate it through a multivariate factorial binomial distribution, whose nature is also given in advance (see Footnote 2).

While other post-PM methods such as 
the Information Bottleneck Method~\cite{tishby2000bottle}
based on rate distortion theory~\cite{davisson1972,cover2012},
Variational Autoencoders~\cite{vae2013,vae2014}, 
Noise-Contrastive Estimation~\cite{nce2010}, 
and Self-Supervised Boosting~\cite{welling2003self} 
also exhibit certain relationships to PM, 
none of them employs gradient-based adversarial NNs in a PM-like minimax game. 
GANs do.

A certain duality between PM variants with attached decoders (Sec. \ref{pmgen})  and GAN variants with 
attached encoders (Sec. \ref{ganfac}) can be illustrated through the following work flow pipelines (view them as very similar 4 step cycles by identifying their beginnings and ends---see Fig. \ref{fig1}):

%\begin{figure}[H]
\begin{figure}[htb]
\begin{center}
\includegraphics[width=\linewidth]{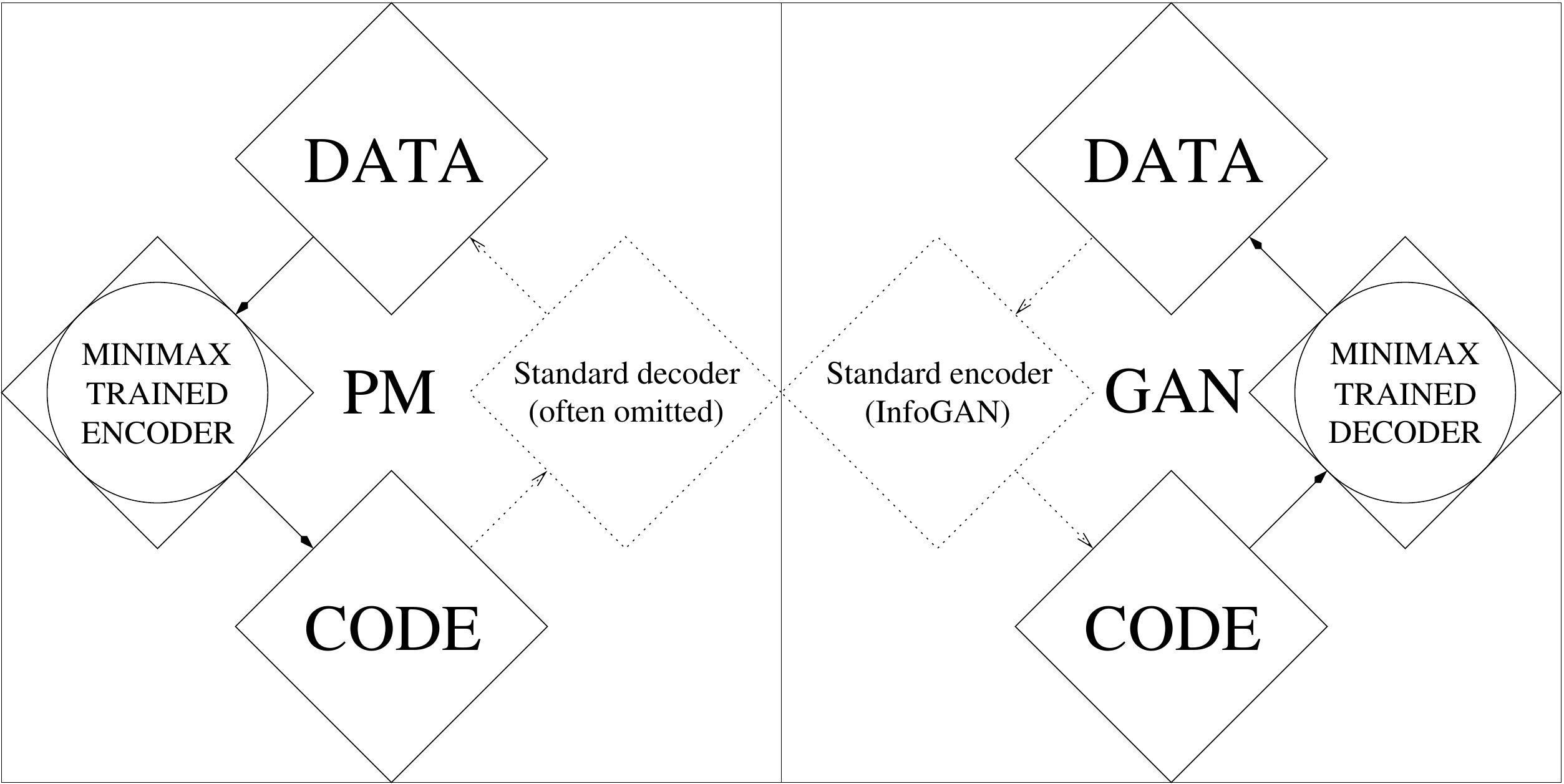}
\end{center}
\caption{{\em
Symmetric work flows of PM and GAN variants. Both PM and GANs model given data distributions in unsupervised fashion. PM uses gradient-based minimax or adversarial training to learn an {\bf en}coder of the data, such that the codes are distributed like the data, and the probability of a given pattern can be read off its code as the product of the predictor-modeled probabilities of the code components (Sec. \ref{pm}). GANs, however, use gradient-based minimax or adversarial training to directly learn a {\bf de}coder of given codes (Sec. \ref{special}). In turn, to decode its codes again, PM can learn a  {\bf non}-adversarial traditional {\bf de}coder (omitted in most PM papers after 1992---see Sec. \ref{pmgen}). Similarly, to encode the data again, GAN variants can learn a  {\bf non}-adversarial traditional {\bf en}coder (absent in the 2014 GAN paper but  compare InfoGANs---see Sec. \ref{ganfac}).
While PM's minimax procedure starts from the data and learns a factorial code in form of a multivariate binomial distribution, GAN's minimax procedure starts from the codes (distributed according to {\em any} user-given distribution), and learns to make data distributed like the original data. 
}}
\label{fig1}
\end{figure}

%\newpage
\begin{itemize}
\item
Pipeline of PM variants with standard decoders:\\
data $\rightarrow$ {\bf minimax-trained encoder} $\rightarrow$  code $\rightarrow$ {\bf traditional decoder} (often omitted) $\rightarrow$  data
\item
Pipeline of GAN variants with standard encoders (compare InfoGANs):\\
code $\rightarrow$ {\bf minimax-trained decoder} $\rightarrow$  data $\rightarrow$ {\bf traditional encoder} $\rightarrow$ code
\end{itemize}

It will be interesting to study experimentally whether the GAN pipeline above is easier to train than PM to make  
factorial codes or useful approximations thereof.

%\newpage
\section{Convergence of Unsupervised Minimax} 
\label{convergence} 

The 2014 GAN paper~\cite{goodfellow2014generative} has a comment on convergence under
the greatly simplifying assumption that one can directly optimize the relevant functions implemented by the two adversaries, without depending on suboptimal local search techniques such as gradient descent. In practice, however, gradient descent is almost always the method of choice. 

So what's really needed is an analysis of what happens when  backpropagation~\cite{Linnainmaa:1970,Linnainmaa:1976,werbos1982sensitivity}
is used for both adversarial networks. Fortunately, there are some relevant results. Convergence can be shown for both GANs and PM through two-time scale stochastic approximation~\cite{Borkar:97,Konda:04,Karmakar:17}.

In fact, Hochreiter's group used this technique to demonstrate convergence for GANs~\cite{Heusel:17arxiv,Heusel:17}; the proof is directly transferrable to the case of PM. 
Of course, such proofs show only convergence to exponentially stable equilibria,
not necessarily to global optima. Compare, e.g.,~\cite{Mazumdar:19}.

%\newpage
\section{Conclusion} 
\label{conclusion} 

The notion of {\em Unsupervised Minimax} refers to unsupervised or self-supervised adaptive modules (typically neural networks or NNs) playing a zero sum game. The first NN somehow learns to generate data. The second NN learns to  predict properties of the generated data, minimizing its error, typically by gradient descent. The first NN maximizes the objective function minimized by the second NN, trying to produce outputs that are hard on the second NN. 
Examples are provided by Adversarial Artificial Curiosity (AC since 1990, Sec. \ref{ac}), Predictability Minimization (PM since 1991, Sec. \ref{pm}), Generative Adversarial Networks (GANs since 2014; conditional GANs since 2010, Sec. \ref{special}). 

This is very different from certain
earlier adversarial machine learning settings
which neither  involved unsupervised NNs nor 
were about modeling  data nor used gradient descent  (see Sec. \ref{intro}, \ref{summary}). 

GANs and cGANs are applications of the AC principle (1990)
where the environment simply returns whether the current output of the first NN is in a given set
 (Sec. \ref{special}). 
 
 GANs are also closely
 related to PM, because both GANs and PM  model the statistics of given data distributions through
gradient-based adversarial nets that play a minimax game (Sec. \ref{pm}). 
%(Unlike AC and GANs, however, PM apparently has never been decribed/used as a generative model as suggested in Sec. \ref{pmgen}.) 
The present paper clarifies some of the previously published confusion surrounding these issues. 

AC's generality (Sec. \ref{general}) extends GAN-like unsupervised minimax to sequential problems, not only for plain pattern generation and classification, but even for RL problems in partially observable environments. In turn, the large body of recent GAN-related insights might help to improve training procedures of 
certain AC systems.

\section*{Acknowledgments} Thanks to Paulo Rauber, Joachim Buhmann, Sepp Hochreiter, Sjoerd van Steenkiste, David Ha, R\'{o}bert Csord\'{a}s, Louis Kirsch, and several anonymous reviewers, for useful comments on a draft of this paper. This work was partially funded by a European Research Council Advanced Grant (ERC no: 742870).

\newpage
\bibliography{bib}
\bibliographystyle{abbrv}
%\bibliographystyle{unsrtnat}
%\bibliographystyle{apalike}
%\bibliographystyle{plain}
%\bibliography{bib,bib_extra}
%\bibliographystyle{alpha}
%\bibliographystyle{apalike}
%\printauthorindex

\end{document}